%% file: bias.tex
\documentclass{article}
\usepackage{nips95}
\usepackage{amsmath}
\usepackage{amsthm}
\usepackage{amssymb}
\usepackage{amsfonts}
\usepackage{mathtools}
\usepackage{epsf}
\input{mathdefs}

\begin{document}

\title{Learning Model Bias}

\author{Jonathan Baxter \\ Department of Computer Science \\
        Royal Holloway College, University of London \\
        {\tt jon@dcs.rhbnc.ac.uk}}
\maketitle

\begin{abstract}
In this paper the problem of {\em learning} appropriate domain-specific bias is 
addressed. It is shown that this can be achieved by learning many related tasks 
from the same domain, and a theorem is given bounding the number tasks that 
must be learnt.
A corollary of the theorem is that if the tasks are known to possess a 
common {\em internal representation} or {\em preprocessing} then
the number of examples required per task for good generalisation when learning
$n$ tasks simultaneously scales like $O(a + \frac{b}{n})$, where 
$O(a)$ is a bound on the minimum 
number of examples requred to learn a single task, and $O(a + b)$ is a bound 
on the number of
examples required to learn each task independently. 
An experiment providing strong qualitative support
for the theoretical results is reported.
\end{abstract}
\section{Introduction}
It has been argued (see \cite{Getal}) that the main
problem in machine learning is the biasing of a learner's hypothesis space
sufficiently well to ensure good generalisation from a small
number of examples. Once suitable biases have been found the actual learning
task is relatively trivial. 
Exisiting methods of bias generally require the input
of a human expert in the form of heuristics, hints \cite{hints2}, domain knowledge,
\etc. Such methods are clearly limited by the accuracy and reliability 
of the expert's knowledge and also by the extent to which that knowledge can be 
transferred to the learner. Here I attempt to solve some of these problems by introducing 
a method for {\em automatically learning} the bias. 

The central idea is that in many learning problems the learner is
typically embedded within an {\em environment} or {\em domain} of related learning
tasks and that the bias appropriate for a single task is likely to be
appropriate for other tasks within the same environment. A simple example is
the problem of handwritten character recognition. A preprocessing stage that 
identifies and removes any (small) rotations, dilations and translations of
an image of a character will be advantageous for recognising all characters.
If the set of all individual character recognition problems is viewed as an
environment of learning tasks, this preprocessor represents a bias that is appropriate
to all tasks in the environment. It is likely that there are many other 
currently unknown biases that are
also appropriate for this environment. We would like to be able to learn these 
automatically.

\begin{figure}
\begin{center}
\leavevmode
\epsfysize=2.0in \epsfbox{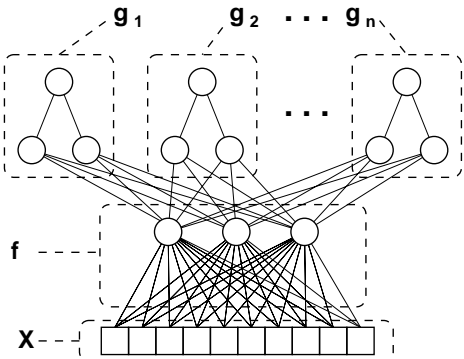}
\caption{Net for learning multiple tasks\label{nnet}. Input $x_{ij}$ from training set $D_i$ 
is propagated forwards through the {\em internal representation} $f$ and then only through
the output network $g_i$. The error $\[g_i(f(x_{ij})) -y_{ij}\]^2$ is similarly 
backpropagated only through the output network $g_i$ and then $f$. Weight updates  
are performed after all training sets $D_1,\dots,D_n$ have been presented.}
\end{center}
\end{figure}

Bias that is appropriate for all tasks must be learnt by sampling 
from many tasks. If only a single task is learnt then the bias extracted is likely to
be specific to that task. For example, if a network is 
constructed as in figure \ref{nnet} and the output nodes are simultaneously trained 
on many similar problems, then the hidden layers are more likely 
to be useful in learning a novel problem of the same type than if only a single problem is
learnt. In the rest of this paper I develop a general theory of bias learning based 
upon the idea of learning multiple related tasks. The theory shows that a learner's
generalisation performance can be greatly improved by learning related tasks and that if
sufficiently many tasks are learnt the learner's bias can be extracted and used to learn
novel tasks. 

Other authors that have empirically investigated the idea of learning 
multiple related tasks include \cite{caruana1} and \cite{thrun}. 
\section{Learning Bias}
\label{theory}
For the sake of argument I consider learning problems that amount to minimizing 
the mean squared error of a function $h$ over some training set $D$.
A more general formulation based on statistical decision theory is 
given in \cite{thesis}. Thus, it is assumed that the learner receives a training set of 
(possibly noisy) {\em input-output} pairs $D=\{(x_1,y_1),\dots,(x_m,y_m)\}$, drawn
according to a probability distribution $P$ on $\XY$ ($X$ being the input space and
$Y$ being the output space) and 
searches through its hypothesis space $\H$ for a function $h\colon X\to Y$ minimizing
the {\em empirical error},
\begin{equation}
\label{emperr}
\Eh(h,D) = \frac1m \sum_{i=1}^m (h(x_i) - y_i)^2.
\end{equation}
The {\em true error} or {\em generalisation error} of $h$ is the expected error under $P$:
\begin{equation}
\label{trueerr}
E(h,P) = \int_{\XY} (h(x) -y)^2\, dP(x,y).
\end{equation}
The hope of course is that an $h$ with a small empirical error on 
a large enough training set
will also have a small true error, \ie it will {\em generalise} well.

I model the {\em environment} of the learner as a pair $(\P,Q)$ where $\P=\{P\}$ is a set 
of learning tasks and $Q$ is a probability measure on $\P$.
The learner is now supplied not with a single hypothesis space $\H$ but with 
a {\em hypothesis space family} $\HH=\{\H\}$. Each $\H\in \HH$ represents a different bias the
learner has about the environment. For example, one $\H$ may contain functions that 
are very smooth, whereas another $\H$ might contain more wiggly functions. Which hypothesis
space is best will depend on the kinds of functions in the environment. 
To determine the best $\H\in \HH$ for $(\P,Q)$,
we provide the learner not with a single training set $D$ but with $n$ such training sets
$D_1,\dots,D_n$. Each $D_i$ is 
generated by first sampling from $\P$ according to $Q$ to give $P_i$ and then sampling
$m$ times from $\XY$ according to $P_i$ to give 
$D_i=\{(x_{i1},y_{i1}),\dots,(x_{im},y_{im})\}$. The learner 
searches for the hypothesis space 
$\H\in \HH$ with minimal empirical error on $D_1,\dots,D_n$, where this is defined by
\begin{equation}
\label{nemperr}
\Eh^*(\H,D_1,\dots,D_n) = \frac1n\sum_{i=1}^n \inf_{h\in\H} \Eh(h,D_i).
\end{equation}
The hypothesis space $\H$ with smallest empirical error is
the one that is best able to learn the $n$ data sets on average.

There are {\em two} ways of measuring the true error of a bias learner. The first is
how well it generalises on the $n$ tasks $P_1,\dots,P_n$ used to generate the training 
sets. Assuming that in the process of minimising \eqref{nemperr} the learner generates
$n$ functions $h_1,\dots,h_n\in\H$ with minimal empirical error on their respective training
sets\footnote{This assumes the infimum in \eqref{nemperr} is attained.}, the learner's
true error is measured by:
\begin{equation}
\label{truenerr1}
E^n(h_1,\dots,h_n,P_1,\dots,P_n) = \frac1n\sum_{i=1}^n E(h_i,P_i).
\end{equation}
Note that in this case the learner's empirical error is given by
$
\Eh^n(h_1,\dots,h_n,D_1,\dots,D_n) = \frac1n\sum_{i=1}^n \Eh(h_i,D_i).
$
The second way of measuring the generalisation error of a bias learner 
is to determine how good $\H$ is for learning {\em novel tasks} 
drawn from the environment $(\P,Q)$:
\begin{equation}
\label{truenerr}
E^*(\H,Q) = \int_\P \inf_{h\in\H} E(h,P)\, dQ(P)
\end{equation} 
A learner that has found an $\H$ with a small value of
\eqref{truenerr} can be said to have {\em learnt to learn} the tasks in $\P$ in
general. To state the bounds ensuring these two types of generalisation a few 
more definitions must be introduced.
\begin{defn}
\label{yuck}
Let $\HH=\{\H\}$ be a hypothesis space family. Let $\HH_\sigma=\{h\in\H\colon \H\in \HH\}$. 
For any $h\colon X\to Y$, define a map $h\colon \XY\to [0,1]$ by $h(x,y) = 
(h(x)-y)^2$. Note the abuse of notation: $h$ stands for two different functions depending
on its argument. 
Given a sequence of $n$ functions $\hv = (h_1,\dots,h_n)$ let $\hv\colon (\XY)^n \to [0,1]$ 
be the function $(x_1,y_1,\dots,x_n,y_n)\mapsto \frac1n\sum_{i=1}^n h_i(x_i,y_i)$. Let 
$\H^n$ be the set of all such functions where the $h_i$ are all chosen from $\H$. Let
$\HH^n = \{\H^n\colon \H\in H\}$. 
For each $\H\in \HH$ define $\H^*\colon \P\to [0,1]$ by 
$\H^*(P) = \inf_{h\in\H} E(h,P)$
and let $\HH^*=\{\H^*\colon\H \in \HH\}$.
\end{defn}
\begin{defn}
\label{metricdef}
Given a set of functions $\H$ from any space $Z$ to $[0,1]$, and any probability measure
on $Z$, define the pseudo-metric $d_P$ on $\H$ by
$$
d_P(h,h') = \int_Z |h(z)-h'(z)|\, dP(z).
$$
Denote the smallest $\ep$-cover of $\(\H,d_P\)$
by $\N\(\ep,\H,d_P\)$. Define the $\ep$-capacity of $\H$ by
$$
\C(\ep, \H) = \sup_P \N\(\ep,\H,d_P\)
$$
where the supremum is over all discrete probability measures $P$ on $Z$.
\end{defn}
Definition \ref{metricdef} will 
be used to define the $\ep$-capacity of spaces such as $\HH^*$ 
and $\Hns$, where from definition \ref{yuck} the latter is 
$\Hns = \{\hv\in\H^n\colon \H\in H\}$.

The following theorem bounds the number of tasks and examples per task required to ensure
that the hypothesis space learnt by a bias learner will, with high probability, contain 
good solutions to novel tasks in the same environment\footnote{The bounds in theorem
\ref{genthm} can be improved to $O\(\frac1\ep\)$ if all $\H\in H$ are convex and the
error is the squared loss \cite{weesun}.}.
\begin{thm}
\label{genthm}
Let the $n$ training sets $D_1,\dots,D_n$ be generated by sampling $n$
times from the environment $\P$ according to $Q$ to give
$P_1,\dots,P_n$, and then sampling $m$ times from each $P_i$ to
generate $D_i$. Let $\HH=\{\H\}$ be a hypothesis space family and
suppose a learner chooses $\hat{\H}\in \HH$ minimizing \eqref{nemperr} on
$D_1,\dots,D_n$. For all $\ep>0$ and $0<\delta<1$, if
\begin{eqnarray*}
n &=& O\(\frac{1}{\ep^2}\ln\frac{\C\(\ep,\HH^*\)}\delta\), \\
\text{and} \quad m &=& O\(\frac1{n\ep^2}\ln\frac{\C(\ep,\Hns)}\delta\)
\end{eqnarray*}
then
$$
\Pr\left\{D_1,\dots,D_n\colon |\Eh^*(\hat{\H},D_1,\dots,D_n) - 
E^*(\hat{\H},Q)| > \ep\right\}
\leq \delta.
$$
\end{thm}
The bound on $m$ in theorem \ref{genthm} 
is the also the number of examples required per task to ensure 
generalisation of the first kind mentioned above. That is, it is the number of examples 
required in each data set $D_i$ to ensure good generalisation on average across all $n$ 
tasks when using the hypothesis space family $\HH$. 
If we let $m(\HH,n,\ep,\delta)$ be
the number of examples
required per task to ensure that 
$\Pr\left\{D_1,\dots,D_n\colon |\Eh^n(h_1,\dots,h_n,D_1,\dots,D_n) - 
E^n(h_1,\dots,h_n,P_1,\dots,P_n)| > \ep\right\}\leq \delta$, where all $h_i\in\H$ for some 
fixed $\H\in \HH$, then 
$$
G(\HH,n,\ep,\delta) = \frac{m(\HH,1,\ep,\delta)}{m(\HH,n,\ep,\delta)}
$$
represents the advantage in learning $n$ tasks as opposed to one task (the ordinary 
learning scenario). Call $G(\HH,n,\ep,\delta)$ the {\em $n$-task gain} of $\HH$.
Using the fact \cite{thesis} that 
$$
\C\(\ep,\HH_\sigma\) \leq \C\(\ep,\Hns\) \leq \C\(\ep,\HH_\sigma\)^n,
$$
and the formula for $m$ from theorem \ref{genthm}, we have,
$$
1\leq G(\HH,n,\ep,\delta) \leq n.
$$
Thus, at least in the worst case analysis here, learning $n$ tasks in the same environment
can result in anything from no gain at all to an $n$-fold reduction in the number of 
examples required per task. In the next section a very intuitive analysis of the
conditions leading to the extreme values of $G(H,n,\ep,\delta)$ is given for the 
situation where an internal representation is being learnt for the environment. I will 
also say more about the bound on the number of tasks ($n$) in theorem \ref{genthm}.
\section{Learning Internal Representations with Neural Networks}
In figure \ref{nnet} $n$ tasks are being learnt using a common representation $f$. In
this case $\Hns$ is the set of all possible networks formed by choosing the weights in the 
representation and output networks.
$\HH_\sigma$ is the same space with a single
output node. If the $n$ tasks were learnt independently (\ie without a common 
representation) then each task would use its own copy of $H_\sigma$, \ie we wouldn't
be forcing the tasks to all use the same representation.

Let $W_R$ be the total number of weights in the representation network and $W_O$ be the
number of weights in an individual output network. Suppose also that all the nodes in each
network are {\em Lipschitz bounded}\footnote{A node $a:\R^p\to \R$ is {\em Lipschitz bounded}
if there exists a constant $c$ such that $|a(x) - a(x')| < c\|x - x'\|$ for all $x,x'\in \R^p$.
Note that this rules out threshold nodes, but sigmoid squashing functions are okay as long
as the weights are bounded.}. Then it can be shown \cite{thesis} that 
$\ln\C\(\ep,\Hns\) = O\(\(W_O + \frac{W_R}n\)\ln\frac1\ep\)$ and 
$\ln\C\(\ep,\HH^*\) = O\(W_R\ln\frac1\ep\)$.
Substituting these bounds into theorem \ref{genthm} shows that to generalise
well on average on $n$ tasks using a common representation requires
$
m = O\(\frac1{\ep^2}\[\(W_O+\frac{W_R}{n}\)\ln\frac1\ep +\frac1n\ln\frac1\delta\]\)
= O\(a + \frac{b}{n}\)
$
examples of each task. In addition, if
$
n = O\(\frac1{\ep^2}W_R\ln\frac1\ep\)
$
then with high probability the resulting representation will be good for learning 
novel tasks from the same environment. Note that this bound is very large. However it
results from a worst-case analysis and so is highly likely to be beaten in practice. 
This is certainly borne out by the experiment in the next section.

The {\em learning gain} $G(H,n,\ep)$ satisfies 
$
G(H,n,\ep) \approx \frac{W_O + W_R}{W_O+\frac{W_R}{n}}.
$
Thus, if $W_R\gg W_O$, $G \approx n$, while if $W_O \gg W_R$ then $G\approx 1$. This 
is perfectly intuitive: when $W_O\gg W_R$ the representation network is hardly doing 
any work, most of the power of the network is in the ouput networks and hence the tasks
are effectively being learnt independently. However, if $W_R\gg W_O$ then the 
representation network dominates; there is very little extra learning to be done for
the individual tasks once the representation is known, and so each example from every 
task is providing full information to the representation network. Hence the gain of $n$.

Note that once a representation has been learnt the sampling burden for learning a novel
task will be reduced to $m=O\(\frac1{\ep^2}\[W_O\ln\frac1\ep + \ln\frac1\delta\]\)$ because
only the output network has to be learnt.
If this theory applies to human learning then the fact that we are able to learn
words, faces, characters, \etc with relatively few examples (a single example in the 
case of faces) indicates that our ``output networks'' are very small, and, given our large
ignorance concerning an appropriate representation, the representation network
for learning in these domains would have to be large, so we would expect to see an 
$n$-task gain of nearly $n$ for learning within these domains.
\section{Experiment: Learning Symmetric Boolean Functions}
\label{expsec}
In this section the results of an 
experiment are reported in which a neural network 
was trained to learn {\em symmetric}\footnote{A symmetric Boolean function is one that 
is invariant under interchange of its inputs, or equivalently, one that only
depends on the number of ``1's'' in its input (\eg parity).}
Boolean functions. The network was the same as the one in figure \ref{nnet} except that
the output networks $g_i$ had no hidden layers.
The input space $X=\{0,1\}^{10}$ was restricted to include only those inputs with between
one and four ones. 
The functions in the environment of the network consisted of
all possible {\em symmetric}  Boolean functions over the 
input space, except the trivial ``constant 0'' and ``constant 1''
functions. 
Training sets $D_1,\dots,D_n$ were generated
by first choosing $n$ functions (with replacement) uniformly from
the fourteen possible, and then choosing $m$ input vectors by choosing a random 
number between 1 and 4 and placing that many 1's at random in the input vector.
The training sets were learnt by minimising the empirical error \eqref{nemperr} using 
the backpropagation algorithm as outlined in figure \ref{nnet}.
Separate simulations were performed with $n$ ranging 
from $1$ to $21$ in steps of four and $m$ ranging from $1$ to $171$ in steps of
$10$. 
Further details of the experimental procedure may be found in
\cite{thesis}, chapter 4.

\begin{figure}[htb]
\begin{center}
\leavevmode
\epsfxsize=2.5in \epsfbox{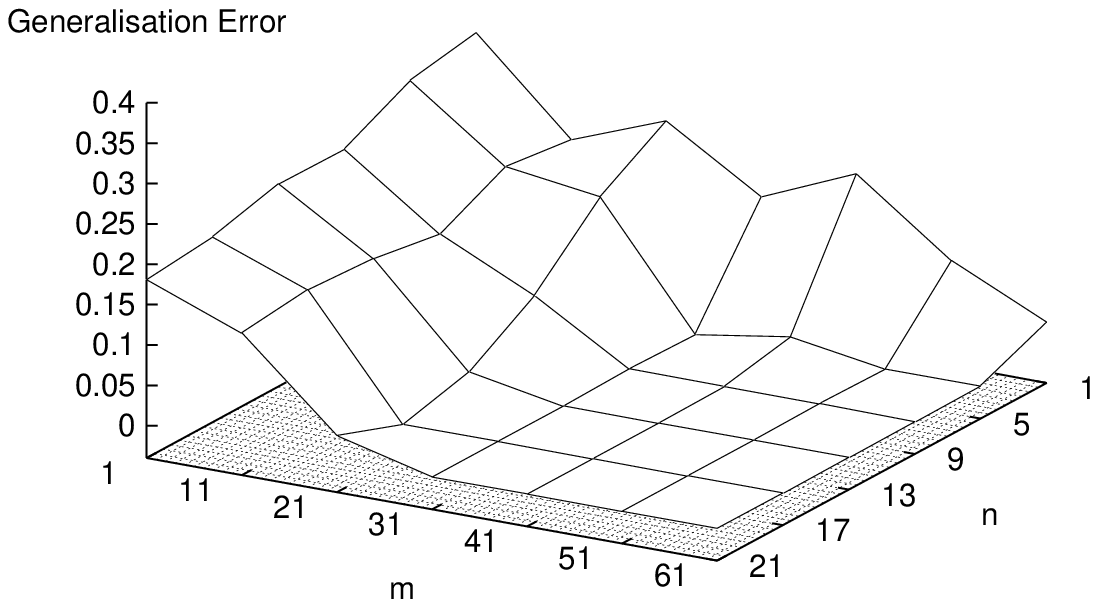}
\epsfxsize=2.5in \epsfbox{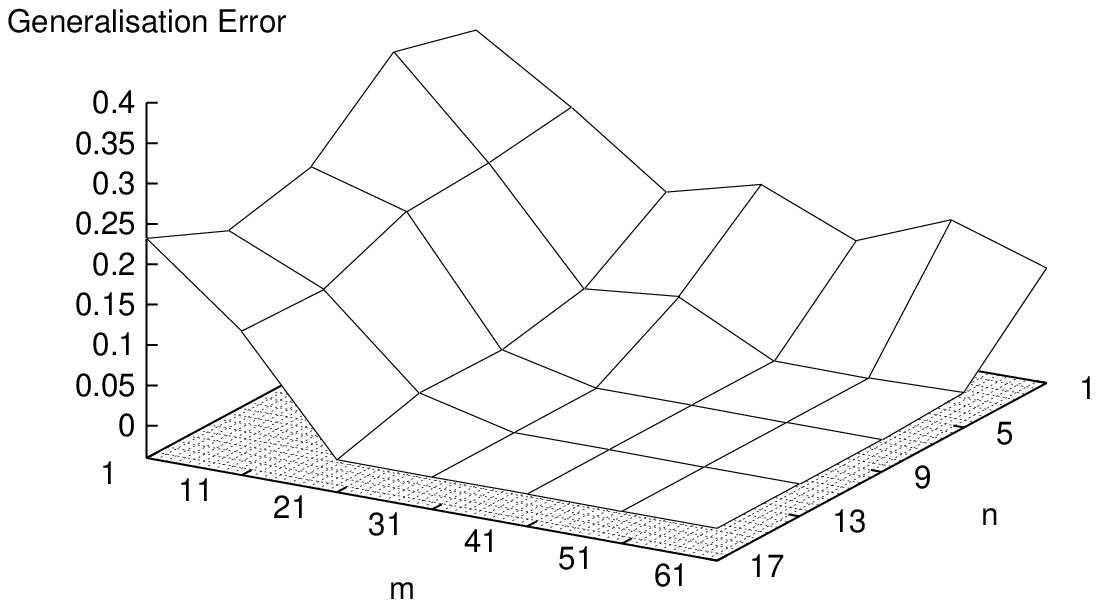}
\caption{Learning {\em surfaces} for two independent simulations.\label{tplot}}
\end{center}
\end{figure}

Once the network had sucessfully learnt the $n$ training sets its generalization
ability was tested on all $n$ functions used to generate the training set.
In this case the generalisation
error (equation \eqref{truenerr1}) could be computed exactly by calculating the
network's output (for all $n$ functions) for each of the $385$ input vectors.
The generalisation error as a function of $n$ and $m$ is plotted in 
figure \ref{tplot} for two independent sets of simulations. 
Both simulations support the theoretical 
result that the number of examples $m$ required for good generalisation
decreases with increasing $n$ ({\em cf} theorem \ref{genthm}). 
For training sets $D_1,\dots,D_n$ that led to a generalisation error of less
than  $0.01$, the representation network $f$ was extracted and tested for 
its {\em true error}, where this is defined as in
equation \eqref{truenerr} (the hypothesis space $\H$ is the set of all networks formed by 
attaching any output network to the fixed representation network $f$).
Although there is insufficient space to show the representation 
error here (see \cite{thesis} for the details), it was found that the representation
error monotonically decreased with the number of tasks learnt, verifying the 
theoretical conclusions.

\begin{figure}[htb]
\begin{center}
\leavevmode
\epsfxsize=1.66in \epsfbox{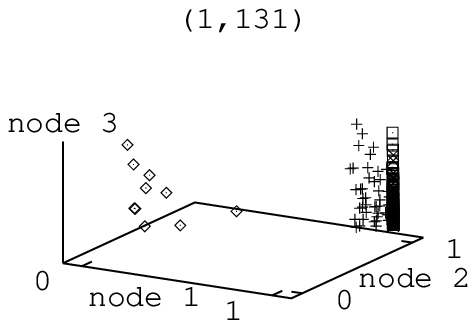}
\epsfxsize=1.66in \epsfbox{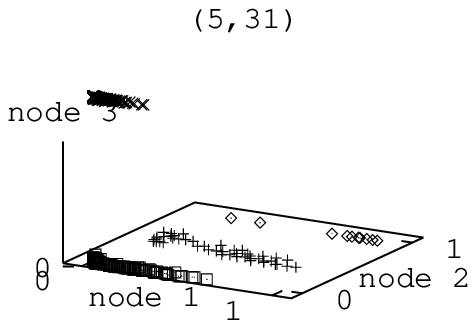}
\epsfxsize=1.66in \epsfbox{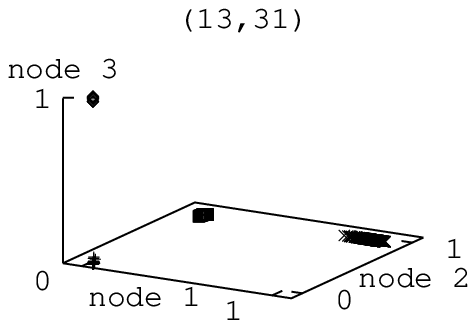}
\caption{Plots of the output of a representation generated from
the indicated $(n,m)$ sample. \label{outplot}}
\end{center}
\end{figure}

The representation's output for all inputs is shown in figure \ref{outplot}
for sample sizes $(n,m) = (1,131),(5,31)$ and $(13,31)$ .
All outputs corresponding to inputs from the same category (\ie the same number of ones) 
are labelled with the same symbol.
The network in the $n=1$ case generalised perfectly but
the resulting representation does not capture
the symmetry in the environment and also does not distinguish the inputs with 
2, 3 and 4 ``1's''
(because the function learnt didn't), showing that learning a single function is not
sufficient to learn an appropriate representation. By $n=5$ the representation's behaviour has
improved (the inputs with differing numbers of 1's are now well separated, but they are still
spread around a lot) and by $n=13$ it is perfect. 
As well as reducing the sampling burden for the $n$ tasks in the training
set, a representation learnt on sufficiently many tasks should be good for
learning novel tasks and should greatly reduce the number of examples
required for new tasks. 
This too was experimentally verified although there is insufficient
space to present the results here (see \cite{thesis}). 
\section{Conclusion}
I have introduced a formal model of bias learning and 
shown that (under mild restrictions) a learner can sample sufficiently 
many times from sufficiently many tasks to learn bias that is appropriate 
for the entire environment. In addition, the number of examples required 
per task to learn $n$ tasks independently was shown to be upper bounded 
by $O(a + b/n)$ for appropriate environments. 
See \cite{Bayes} for an analysis of bias learning within an Information theoretic 
framework which leads to an exact $a + b/n$-type bound.
\vspace*{-5mm}
\bibliographystyle{abbrv}
\bibliography{bib}
\end{document}

%% file: mathdefs.tex
\newcommand{\bb}{\mathbb}

\renewcommand{\glossary}[2]{}
\def\addcontentsline#1#2#3{} 

\newtheorem{thm}{Theorem}

\newtheorem{defn}{Definition}

\newcommand{\ie}{{\em i.e.\ }}

\newcommand{\eg}{{\em e.g.\ }}

\newcommand{\etc}{{\em etc\ }}

\newcommand{\HH}{{\bb H}}

\renewcommand{\H}{{\cal H}}
\newcommand{\XY}{{X\times Y}}

\renewcommand{\P}{{\cal P}}

\newcommand{\Eh}{{\hat E}}

\newcommand{\be}{\begin{equation*}}

\newcommand{\ep}{{\varepsilon}}

\newcommand{\C}{{\cal C}}
\newcommand{\N}{{\cal N}}

\renewcommand{\(}{\left(}
\renewcommand{\)}{\right)}
\renewcommand{\[}{\left[}
\renewcommand{\]}{\right]}

\renewcommand{\hbar}{{\overline h}}

\newcommand{\hv}{{\vec h}}

\newcommand{\R}{{\mathbb R}}

\newcommand{\Hns}{{\[\HH^n\]_\sigma}}